\documentclass[journal,transmag]{IEEEtran}

\usepackage{times}
\usepackage{epsfig}
\usepackage{graphicx, subfigure}
\usepackage{lscape}

\usepackage{graphicx}
\usepackage{amsmath}
\usepackage{amssymb}
\usepackage{amsthm}
\usepackage{cite}
\usepackage{url}
\usepackage{color}
\usepackage{tikz}
\usepackage{threeparttable}
\usepackage{multirow}

\usepackage{makeidx, tabularx, graphicx, setspace, amsmath, longtable, multirow, float, theorem, times, booktabs, siunitx, lineno}

\definecolor{Red}{rgb}{1, 0.2, 0.2}

\newcommand{\zzz}[1]{\textcolor{black}{#1}}

\ifCLASSINFOpdf

\else

\fi

\begin{document}

%
\title{Convolutional Invasion and Expansion Networks \\ for Tumor Growth Prediction
\\
}


\author{\IEEEauthorblockN{Ling Zhang\IEEEauthorrefmark{},
Le Lu\IEEEauthorrefmark{},~\IEEEmembership{Senior Member,~IEEE},
Ronald M. Summers\IEEEauthorrefmark{}, 
Electron Kebebew\IEEEauthorrefmark{}, and\\ 
Jianhua Yao\IEEEauthorrefmark{}}

\thanks{This work was supported by the Intramural Research Program at the NIH Clinical Center. \emph{Corresponding author: Jianhua Yao.}}
\thanks{L. Zhang, L. Lu, R. M. Summers, and J. Yao are with the Imaging Biomarkers and Computer-Aided Diagnosis Laboratory and the Clinical Image Processing Service, Radiology and Imaging Sciences Department, National Institutes of Health Clinical Center, Bethesda, MD 20892, USA (e-mail: ling.zhang3@nih.gov; jyao@cc.nih.gov).}
\thanks{E. Kebebew is with the Endocrine Oncology Branch, National Cancer Institute, National Institutes of Health, Bethesda, MD 20892, USA.}
}


\maketitle

\begin{abstract}
Tumor growth is associated with cell invasion and mass-effect, which are traditionally formulated by mathematical models, namely reaction-diffusion equations and biomechanics. Such models can be personalized based on clinical measurements to build the predictive models for tumor growth. In this paper, we investigate the possibility of using deep convolutional neural networks (ConvNets) to directly represent and learn the cell invasion and mass-effect, and to predict the subsequent involvement regions of a tumor. The invasion network learns the cell invasion from information related to metabolic rate, cell density and tumor boundary derived from multimodal imaging data. The expansion network models the mass-effect from the growing motion of tumor mass. We also study different architectures that fuse the invasion and expansion networks, in order to exploit the inherent correlations among them. Our network can easily be trained on population data and personalized to a target patient, unlike most previous mathematical modeling methods that fail to incorporate population data. Quantitative experiments on a pancreatic tumor data set show that the proposed method substantially outperforms a state-of-the-art mathematical model-based approach in both accuracy and efficiency, and that the information captured by each of the two subnetworks are complementary.
\end{abstract}



\maketitle

\IEEEdisplaynontitleabstractindextext

\IEEEpeerreviewmaketitle


\section{Introduction}

\begin{figure}[!t]
   \begin{center}
   \begin{tabular}{c}
   \includegraphics[width=7.5cm]{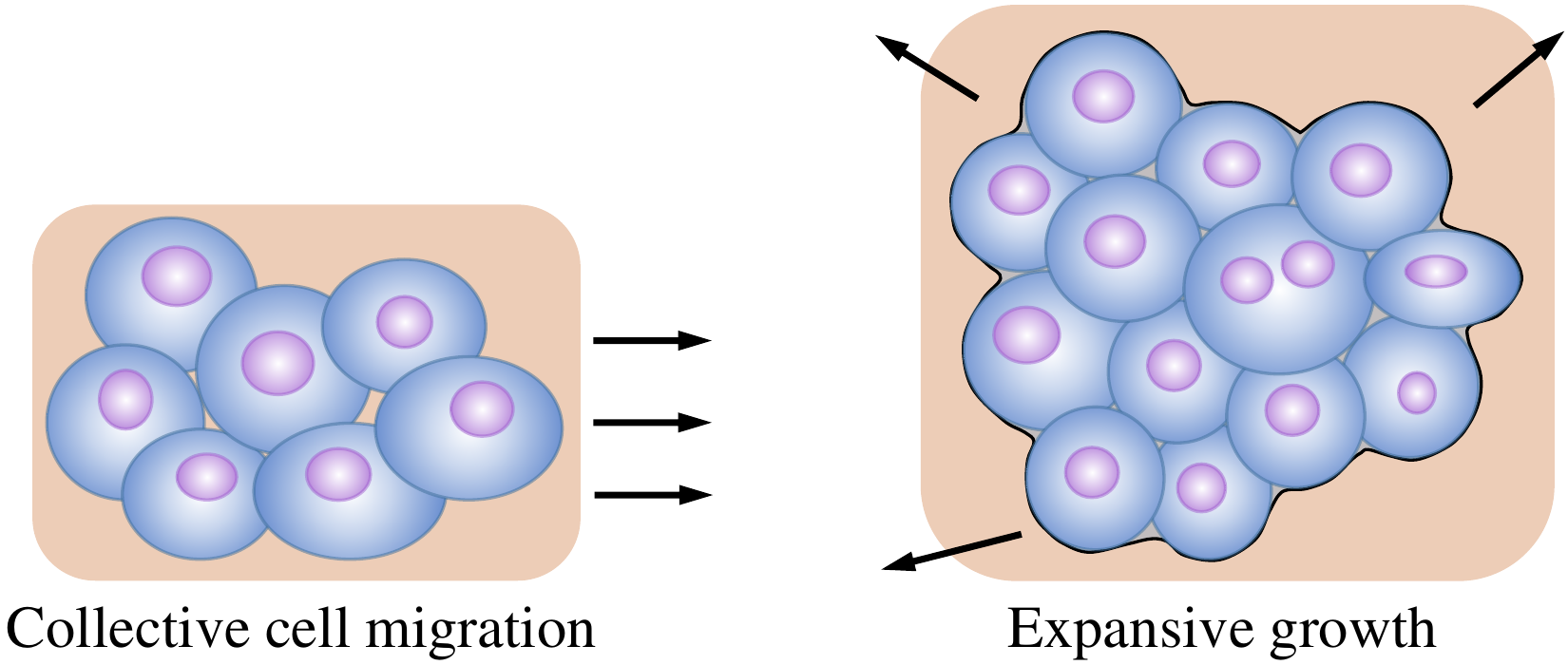}
   \end{tabular}
   \end{center}
   \caption[example] 
   { \label{figbio} 
The two fundamental processes of tumor growth: cell invasion and expansive growth of tumor cells. 
}
   \end{figure}

Cancer cells originate from the irreversible injuring of respiration of normal cells. Part of the injured cells could succeed in replacing the lost respiration energy by fermentation energy, but will therefore convert into undifferentiated and widely growing cells (cancer cells) \cite{warburg1956origin}. Tumors develop from such abnormal cell/tissue growth, which is associated with cell invasion and mass-effect \cite{friedl2012classifying}. Cell invasion is characterized by the migration and penetration of cohesive groups of tumor cells to surrounding tissues, and mass-effect by the distension and outward pushing of tissues induced by tumor growth (Fig. \ref{figbio}). 

\zzz{
Medical imaging data provides non-invasive and in vivo measurements of the tumor morphology and underlying tumor physiological parameters over time. For example, dual phase contrast-enhanced CT is the most readily available modality for evaluation of tumor morphology and cell density in clinical environments; In recent years, FDG-PET (2-[18F] Fluoro-2-deoxyglucose positron emission tomography) and MRI are gaining popularity in characterizing different tumor properties, such as metabolic rate and fluid characteristics \cite{keutgen2016evaluation,liu2014patient,wong2015tumor,wong2017pancreatic}. 
For tumor growth assessment, RECIST (Response Evaluation Criteria in Solid Tumors), where the longest diameter of a tumor is measured \cite{weisbrod2014assessment}, is the current standard of practice. RECIST has its limitation since it is only one dimensional measurement. Mathematical modeling, which represents the tumor growth process as a physiological and biomechanical model and personalizes the model based on clinical measurements of a target patient, can predict the entire tumor volume including its size, shape and involved region.
Therefore, data-driven tumor growth modeling has been actively studied \cite{swanson2000quantitative,clatz2005realistic,hogea2007modeling,hogea2008image,menze2011generative,chen2013kidney,liu2014patient,wong2015tumor,wong2017pancreatic}.}

In most previous model\zzz{-based methods} \cite{clatz2005realistic,hogea2008image,chen2013kidney,liu2014patient,wong2015tumor,wong2017pancreatic}, both cell invasion and mass-effect are accounted for, since they are inter-related, mutually reinforcing factors \cite{friedl2012classifying}. Cell invasion is often modeled by the reaction-diffusion equations \cite{swanson2000quantitative,clatz2005realistic,hogea2008image,menze2011generative,chen2013kidney,liu2014patient,wong2015tumor,wong2017pancreatic}, and mass-effect by the properties of passive material (mainly isotropic materials) and active growth (biomechanical model) \cite{clatz2005realistic,hogea2007modeling,hogea2008image,chen2013kidney,liu2014patient,wong2015tumor,wong2017pancreatic}. 
While these methods yield informative results, most previous tumor growth models are independently estimated from the target patient without considering the tumor growth pattern of population trend. Furthermore, the small number of model parameters (e.g., 5 in \cite{wong2017pancreatic}) may be insufficient to represent the complex characteristics of tumor growth. 

Apart from these mathematical modeling methods, a different idea based on voxel motion is proposed \cite{weizman2012prediction}. By computing the optical flow of voxels over time, and estimating the future deformable field via an autoregressive model, this method is able to predict entire brain MR scan. However, the tumor growth pattern of population trend is still not involved. Moreover, this method might over-simplify the tumor growth process, since it infers the future growth in a linear manner (most tumor growth are nonlinear).

Data-driven statistical learning is a potential solution to incorporate the population trend of tumor growth into personalized tumor modeling. The pioneer study in \cite{morris2006learning} attempts to model the glioma growth patterns as a classification problem. This model learns tumor growth patterns from selected features at patient, tumor, and voxel levels, and achieves a prediction accuracy (both precision and recall) of 59.8\%. However, this study only learns population trend of tumor growth without incorporating subject-specific personalization related to the tumor natural history. Besides this problem, this early study is limited by the feature design and selection components. Specifically, hand-crafted features are extracted to describe each isolated voxel (without context information). These features could be compromised by the limited understanding of tumor growth, and some of them are obtained in an unsupervised manner. Furthermore, some features may not be generally effective for other tumors, e.g., the tissue type features (cerebrospinal fluid, white and grey matter) in brain tumors \cite{morris2006learning} are not fit for liver or pancreatic tumors. Moreover, considering that the prediction of tumor growth pattern is challenging even for human experts, the low-level features used in this study may not be able to represent complex discriminative information. 

Deep neural networks \cite{lecun2015deep} are high capacity trainable models with a large set of ($\sim$ 15 M) parameters. By optimizing the massive amount of network parameters using gradient backpropagation, the network can discover and represent intricate structures from raw data without any type of feature-engineering. In particular, deep convolutional neural networks (ConvNets) \cite{lecun1989backpropagation,krizhevsky2012imagenet} have significantly improved performance in a variety of traditional medical imaging applications \cite{greenspan2016guest}, including lesion detection \cite{holger2016improving}, anatomy segmentation \cite{moeskops2016automatic}, and pathology discrimination \cite{zhang2017deeppap}. The basic idea of these applications is using deep learning to determine the current status of a pixel or an image (whether it belongs to object boundary/region, or certain category). The ConvNets have also been successfully used in prediction of future binary labels at image\zzz{/patient} level, such as survival prediction of patients with brain and lung cancer \cite{nie20163d,yao2016imaging,zhu2017wsisa}. 
\zzz{Another direction of future prediction is on pixel-level, which reconstructs the entire tumor volume, and therefore characterize the size, shape and involved region of a tumor. Moreover, a patient may have a number of tumors and they may have different growth patterns and characteristics. A single prediction for the patient would be ambiguous. In all, pixel-level prediction is more desirable for precision medicine, as it can potentially lead to better treatment management and surgical planning.}
\zzz{In this work, we are investigating} whether deep ConvNets are capable of predicting the future status at the pixel/voxel level for medical problem.

More generally, in computer vision and machine learning community, the problem of modeling spatio-temporal information and predicting the future have attracted lots of research interest in recent years. The spatio-temporal ConvNet models \cite{simonyan2014two,feichtenhofer2016convolutional}, which explicitly represent the spatial and temporal information as RGB raw intensity and optical flow magnitude \cite{brox2004high}, respectively, have shown outstanding performance for action recognition. To deal with the modeling of future status, recurrent neural network (RNN) and ConvNet are two popular methods. RNN has a ``memory" of the history of previous inputs, which can be used to influence the network output \cite{graves2008supervised}. RNN is good at predicting the next word in a sequence \cite{lecun2015deep}, and has been used to predict the next image frames in video \cite{ranzato2014video,srivastava2015unsupervised,finn2016unsupervised}. ConvNet with fully convolutional architecture can also be directly trained to predict next images in video \cite{mathieu2015deep} by feeding previous images to the network. However, the images predicted by both RNN and \zzz{fully} ConvNet are blurry, even after re-parameterizing the problem of predicting raw pixels to predicting pixel motion distribution \cite{finn2016unsupervised}, or improving the predictions by multi-scale architecture and adversarial training \cite{mathieu2015deep}. Actually, directly modeling the future raw intensities might be an over-complicated task \cite{neverova2017predicting}. Therefore, predicting the future high-level object properties, such as object boundary \cite{bhattacharyya2016long} or semantic segmentation \cite{neverova2017predicting}, has been exploited recently. It is also demonstrated in \cite{bhattacharyya2016long} that the \zzz{fully} ConvNet-based method can produce more accurate boundary prediction in compared to the RNN-based method. In addition, \zzz{fully} ConvNet has shown its strong ability to predict the next status at image-pixel level -- as a key component in AlphaGo \cite{maddison2014move,silver2016mastering}, \zzz{fully} ConvNets are trained to predict the next move (position of the $19\times19$ Go game board) of Go player, given the current board status, with an accuracy of 57\%. 

Therefore, in this paper, we investigate whether ConvNets can be used to directly represent and learn the two fundamental processes of tumor growth (cell-invasion and mass-effect) from multi-model tumor imaging data at multiple time points. Moreover, given the current state information in the data, we determine whether the ConvNet is capable of predicting the future state of the tumor growth. 
Our proposed ConvNet architectures are partially inspired by the mixture of policy and value networks for evaluating the next move/position in game of Go \cite{silver2016mastering}, as well as the integration of spatial and temporal networks for effectively recognizing action in videos \cite{simonyan2014two,feichtenhofer2016convolutional}. In addition to $x$ and $y$ direction optical flow magnitudes (i.e., 2-channel image input) used in \cite{simonyan2014two,feichtenhofer2016convolutional}, we add the flow orientation information to form a 3-channel input, as the optical flow orientation is crucial to tumor growth estimation. In addition, we apply a personalization training step to our networks which is necessary and important to patient-specific tumor growth modeling \cite{chen2013kidney,liu2014patient,wong2015tumor,wong2017pancreatic}. Furthermore, we focus on predicting future labels of tumor mask/segmentation, which is found to be substantially better than directly predicting and then segmenting future raw images \cite{neverova2017predicting}. Finally, considering that the longitudinal tumor datasets spanning multiple years are very hard to obtain, 
the issue of small dataset is alleviated by patch oversampling strategy and pixel-wise ConvNet learning (e.g., only a single anatomical MR image is required to train a ConvNet for accurate brain image segmentation \cite{moeskops2016automatic}), in contrast to the fully ConvNet used in \cite{silver2016mastering,bhattacharyya2016long,neverova2017predicting} which is more efficient but may lose labeling accuracy. 

The main contributions of this paper can be summarized as: 1) To the best of our knowledge, this is the first time to use learnable ConvNet models for explicitly capturing these two fundamental processes of tumor growth. 
2) The invasion network can make its prediction based on the metabolic rate, cell density and tumor boundary, all derived from the multi-model imaging data. Mass-effect -- the mechanical force exerted by the growing tumor -- can be approximated by the expansion/shrink motion (magnitude and orientation) of the tumor mass. This expansion/shrink cue is captured by optical flow computing \cite{brox2004high,baker2011database}, based on which the expansion network is trained to infer tumor growth. 3) To exploit the inherent correlations among the invasion and expansion information, we study and evaluate three different network architectures, named: early-fusion, late-fusion, and end-to-end fusion. 4) Our proposed ConvNet architectures can be both trained using population data and personalized to a target patient. Quantitative experiments on a pancreatic tumor dataset demonstrate that the proposed method substantially outperforms a state-of-the-art model-based method \cite{wong2017pancreatic} in both accuracy and efficiency. The new method is also much more efficient than our recently proposed group learning method \cite{zhang2017personalized} while with comparable accuracy.

   \begin{figure}[!t]
   \begin{center}
   \begin{tabular}{c}
   \includegraphics[width=8.5cm]{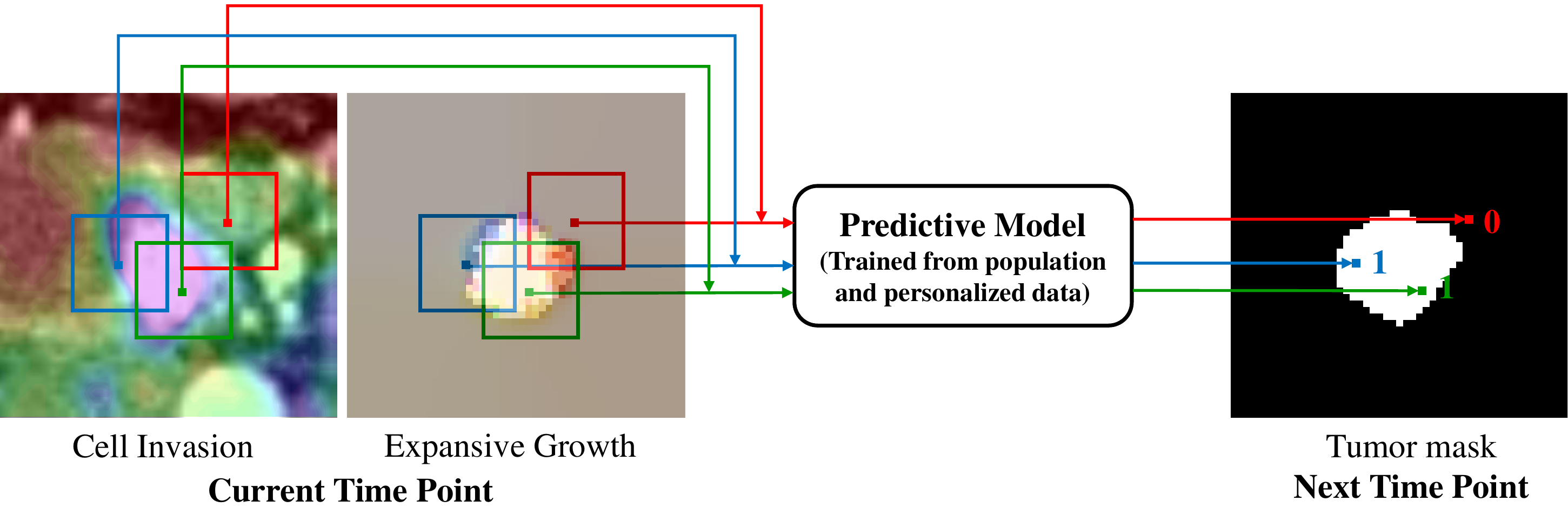}
   \end{tabular}
   \end{center}
   \caption[example] 
   { \label{figidea} 
Basic idea of the voxel-wise prediction of tumor growth based on cell invasion and expansion growth information. 
}
   \end{figure} 

\section{Convolutional Invasion and Expansion Networks}
\label{convIEnets}

The basic idea of our method is using a learned predictive model to predict whether the voxels in current time point will be tumor or not at the next time point, as shown in Fig. \ref{figidea}. The inputs to the predictive model are image patches (sampled around the tumor region) representing cell invasion and expansive growth information that are derived from multimodal imaging data. The corresponding outputs are binary prediction labels: 1 (if the input patch center will be in tumor region at the next time point) or 0 (otherwise). The overview of learning such a predictive model is described below.

   \begin{figure}[!t]
   \begin{center}
   \begin{tabular}{c}
   \includegraphics[width=8.6cm]{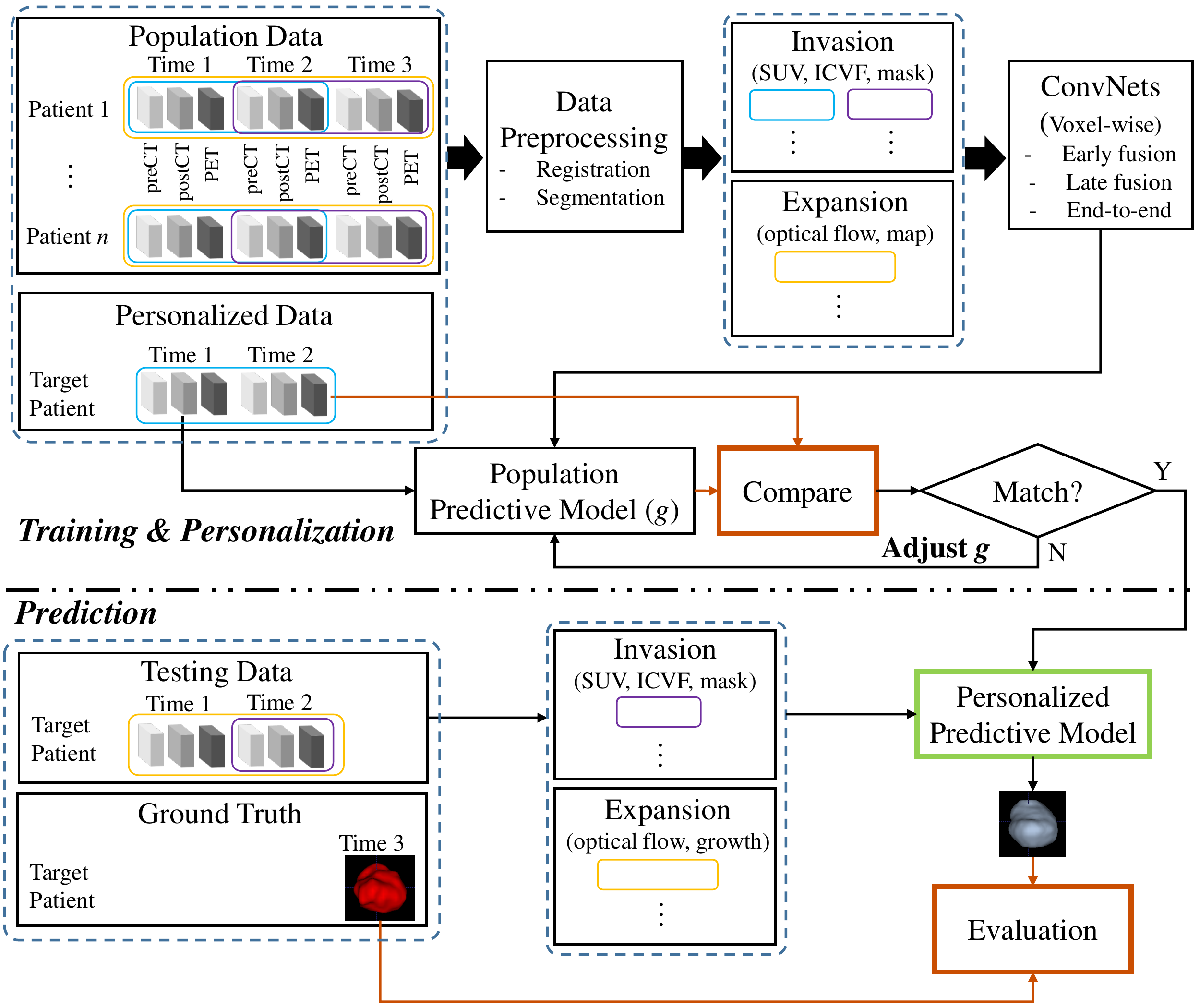}
   \end{tabular}
   \end{center}
   \caption[example] 
   { \label{figframework} 
Overview of the proposed framework for predicting tumor growth. The upper part is model training (learn population trend) \& personalization and the lower part is unseen data prediction. The blue and purple empty boxes indicate the data used for generating invasion information; the yellow empty boxes for expansion information.
}
   \end{figure} 

Particularly for the longitudinal tumor data in this study, every patient has multimodal imaging (dual phase contrast-enhanced CT and FDG-PET) at three time points spanning between three to four years, we design a training \& personalization and prediction framework as illustrated in Fig. \ref{figframework}. The imaging data of different modalities and at different time points are first registered and the tumors are segmented. The intracellular volume fraction (ICVF) and standardized uptake value (SUV) \cite{liu2014patient} are computed. Along with tumor mask, a 3-channel image that reveals both functional and structural information about the tumor's physiological status serve as the input of invasion subnetwork. The input of the expansion subnetwork is a 4-channel image, containing the 3-channel optical flow image \cite{baker2011database} (using a color encoding scheme for flow visualization \cite{baker2011database}) carrying the growing motion, and the growth map of tumor mass across time1 and time2. In the training \& personalization stage, voxel-wise ConvNets are trained from all the pairs of time points (time1/time2, time2/time3, and \zzz{(}time1$\rightarrow$time2\zzz{)}/time3) from population data, and then personalized on pair of time1/time2 from personalized data by adjusting model parameters. \zzz{Note that (time1$\rightarrow$time2) means the expansion data from time1 to time2, and time3 provides data label (future tumor or not).} In the prediction stage, given the data of target patient at time1 and time2, invasion and expansion information are fed into the personalized predictive model to predict the tumor region at a future time point 3 in a voxel-wise manner. It should be pointed out that the training and personalization/test sets are separated at the patient-level, and the testing data (predicting time3 based on time1 and time2 of the target patient) is totally unseen for the predictive model.

\subsection{Learning Invasion Network}
\label{invasionnet}
\subsubsection{Image Processing and Patch Extraction}
\label{invasionpatch}
\begin{figure}[!t]
   \begin{center}
   \begin{tabular}{c}
   \includegraphics[width=8.5cm]{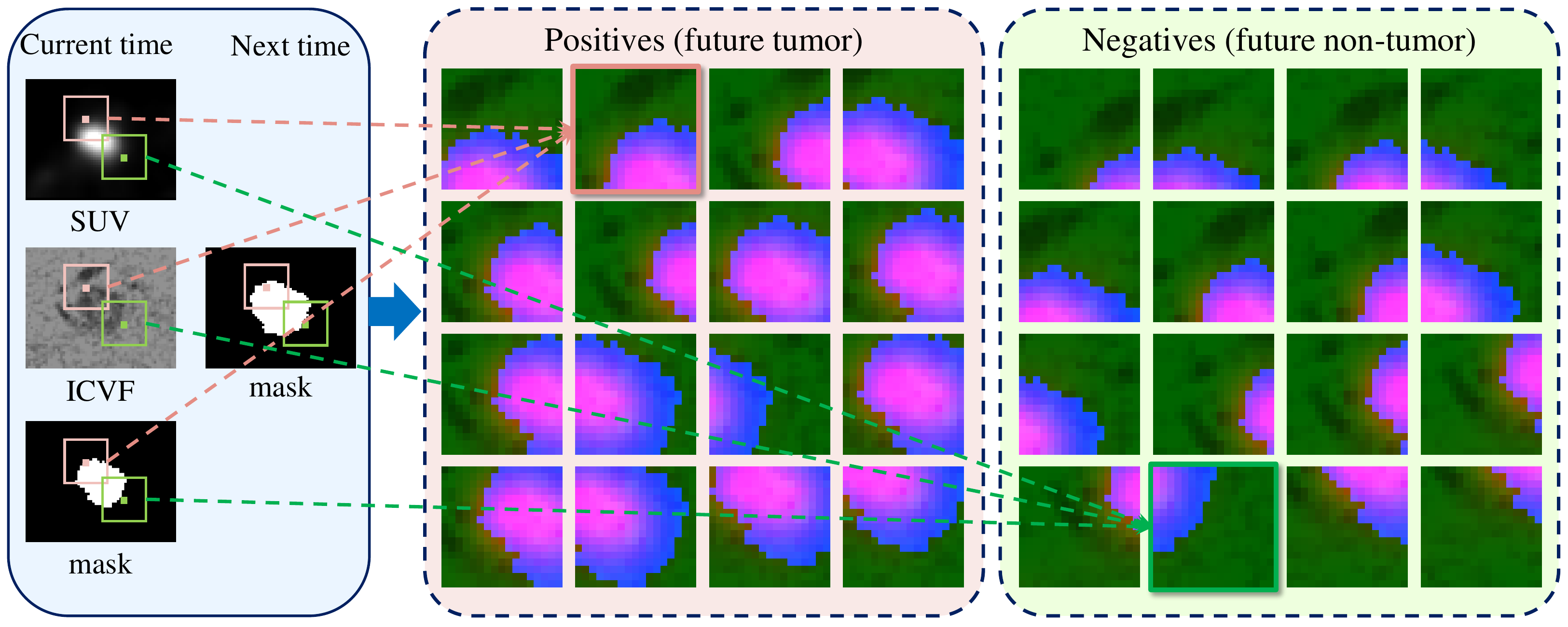}
   \end{tabular}
   \end{center} \vspace{-2mm}
   \caption[example] 
   { \label{figcrop} 
Some examples of positive (center panel) and negative (right panel) training samples. In the left panel, the pink and green bounding boxes at the current time illustrate the cropping of a positive sample and a negative sample from multimodal imaging data. Each sample is a three-channel RGB image formed by the cropped SUV, ICVF, and mask at the current time. The label of each sample is determined by the location of corresponding bounding box center at the next time - inside
tumor (pink): positive; outside tumor (green): negative.
} \vspace{-3mm}
   \end{figure}

To establish the spatial-temporal relationship of tumor growth along different time points, the multi-model imaging data are registered based on mutual information and imaging data at different time points are aligned using the tumor center \cite{wong2017pancreatic}. After that, three types of information (SUV, ICVF, and tumor mask, refer to the left panel in Fig. \ref{figcrop} as an example) related to tumor property are extracted from the multimodal images and used as a three-channel input to the invasion ConvNet model.

(1) The FDG-PET characterizes regions in the body which are more active and need more energy to maintain existing tumor cells and to create new tumor cells. This motivates us to use FDG-PET to measure metabolic rate and incorporate it in learning the tumor predictive model. SUV is a quantitative measurement of the metabolic rate \cite{liu2014patient}. 
To adapt to the ConvNets model, the SUV values from PET images are magnified by 100 followed by a cutting window [100 2600] and then transformed linearly to [0 255].

   \begin{figure*}[!t]
   \begin{center}
   \includegraphics[width=13.8cm]{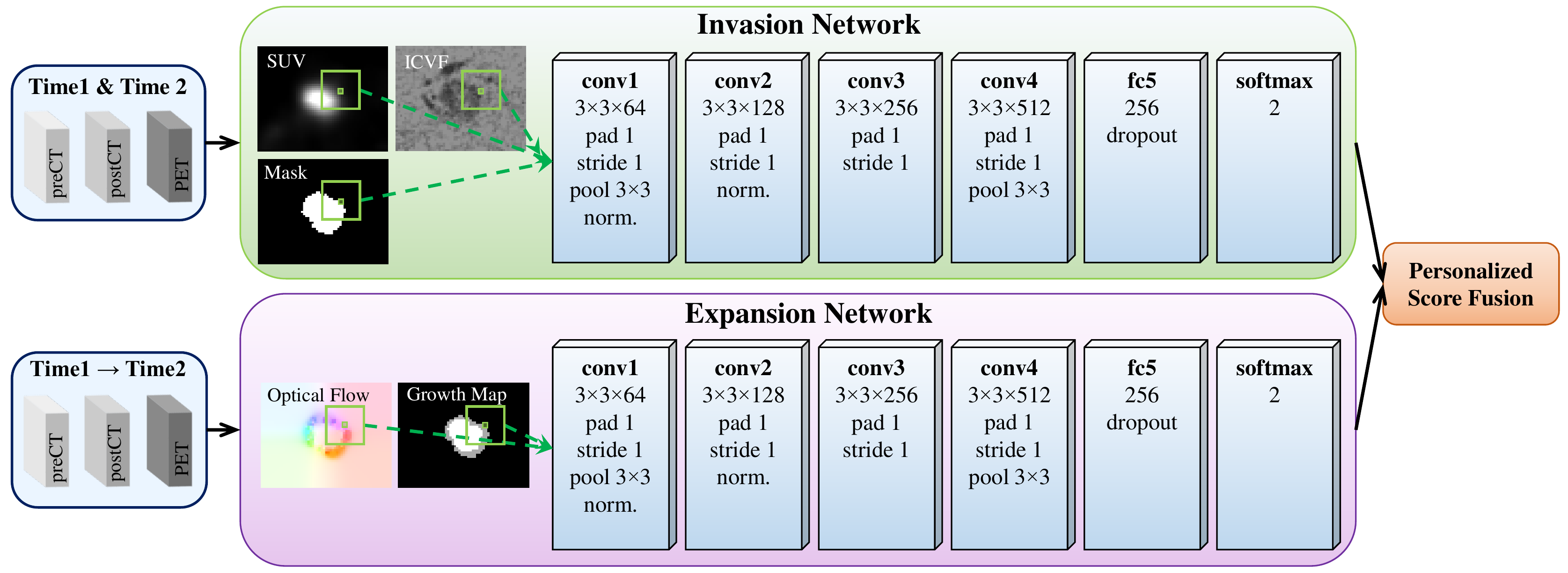}
   \end{center} \vspace{-3mm}
   \caption[example] 
   { \label{figlatefusion}  
ConvNet architecture for late fusion of the invasion and expansion networks for predicting tumor growth.
} \vspace{-3mm}
   \end{figure*} 

(2) Tumor grade is one of the most important prognosticators, and is determined by the proliferation rate of the neoplastic cells \cite{bosman2010classification}. This motivates us to extract the underlying physiological parameter related to the cell number. ICVF is an representation of the normalized tumor cell density, and is computed from the registered dual-phase contrast-enhanced CT:
\begin{equation}
{\rm ICVF} = 1 - \frac{HU_{post\_tumor} - HU_{pre\_tumor}}{E[HU_{post\_blood} - HU_{pre\_blood}]} \times (1 - Hct)
\end{equation}
where $HU_{post\_tumor}$, $HU_{pre\_tumor}$, $HU_{post\_blood}$, and $HU_{pre\_blood}$ are the Hounsfield units of the post- and pre-contrast CT images at the segmented tumor and blood pool (aorta), respectively. $E[\bullet]$ represents the mean value. $Hct$ is the hematocrit which can be obtained from blood samples, thus the ICVF of the tumor is computed using the ICVF of blood ($Hct$) as a reference. The resulting ICVF values are magnified by 100 (range between [0 100]) for ConvNets input.

(3) Tumor stage is another important prognosticator, and is determined by the size and extend of the tumor \cite{bosman2010classification}. Previous studies have used the tumor mask/boundary to monitor the tumor morphological change and estimate model parameters \cite{swanson2000quantitative,clatz2005realistic,hogea2008image}. In this study, \zzz{following \cite{wong2017pancreatic},} the tumors are segmented by a semiautomatic level set algorithm with region competition \cite{neuroimage2006} on the post-contrast CT image to form tumor masks with binary values (0 or 255). 

As illustrated in Fig. \ref{figidea}, to train a ConvNet to distinguish between future tumor and future non-tumor voxels, image patches of size $17\times 17$ voxels $\times 3$ -- centered at voxels near the tumor region at the current time point -- are sampled from four channels of representations reflecting and modeling the tumor's physiological status. Patches centered inside or outside of tumor regions at the next time point are labeled as ``1" and ``0", serving as positive and negative training samples, respectively. This patch based extraction method allows for embedding the context information surrounding the tumor voxel. The voxel (patch center) sampling range is restricted to a bounding box of $\pm 15$ pixels centered at the tumor center, as the pancreatic tumors in our dataset are $ < 3$ cm ($\approx 30$ pixels) in diameter and are slow-growing. To avoid the classification bias towards the majority class (non-tumor) and to improve the accuracy and convergence rate during ConvNet training \cite{krizhevsky2012imagenet,zhang2017deeppap}, we create a roughly balanced training set by proportionally under-sampling the non-tumor patches. A few examples of positive and negative patches of SUV, ICVF, and mask encoded in three-channel RGB color images are shown in Fig. \ref{figcrop}.

\subsubsection{Network Architecture}
\label{invarchitecture}
We use a six-layer ConvNet adapted from AlexNet \cite{krizhevsky2012imagenet}, which includes 4 convolutional ($conv$) layers and 1 fully connected ($fc$) layers (cf. upper panel in Fig. \ref{figlatefusion}).
The inputs are of size $17\times 17\times 3$ image patch stacks, where 3 refers to the tumor status channels of SUV, ICVF, and tumor mask. All $conv$ layer filters are of size $3\times 3$, with padding and stride of 1. The number of filters from $conv1$ to $conv4$ layers are 64, 128, 256, and 512, respectively. Max-pooling is performed over $3\times 3$ spatial windows with stride 2 for $conv1$ and $conv4$ layers. Local response normalization is used for $conv1$ and $conv2$ layers using the same setting as \cite{krizhevsky2012imagenet}. The $fc5$ layer contains 256 rectifier units and applies ``dropout" to reduce overfitting. All layers are equipped with the ReLU (rectified linear unit) activation function. The output layer is composed of two neurons corresponding to the classes future tumor or non-tumor, and applies a softmax loss function. The invasion ConvNet is trained on image patch-label pairs from scratch on all pairs of time points (time1/time2 and time2/time3) from the population dataset.

\subsection{Learning Expansion Network}
\subsubsection{Image Processing and Patch Extraction}
Unlike the invasion network, which performs predictions from static images, the expansion network accounts for image motion information. Its input images, of size $17 \times 17 \times 4$, capture expansion motion information between two time points. 3 channels derive from a color-coded 3-channel optical flow image, and the 4$^{th}$ from a tumor growth map between time1 and time2.  
Such images explicitly describe the past growing trend of tumor mass, as an image-based approximation of the underlying biomechanical force exerted by the growing tumor. These patches are sampled using the same restriction and balancing schemes applied for the invasion network (Section \ref{invasionpatch}).

   \begin{figure*}[!t]
   \begin{center}
   \includegraphics[width=13.8cm]{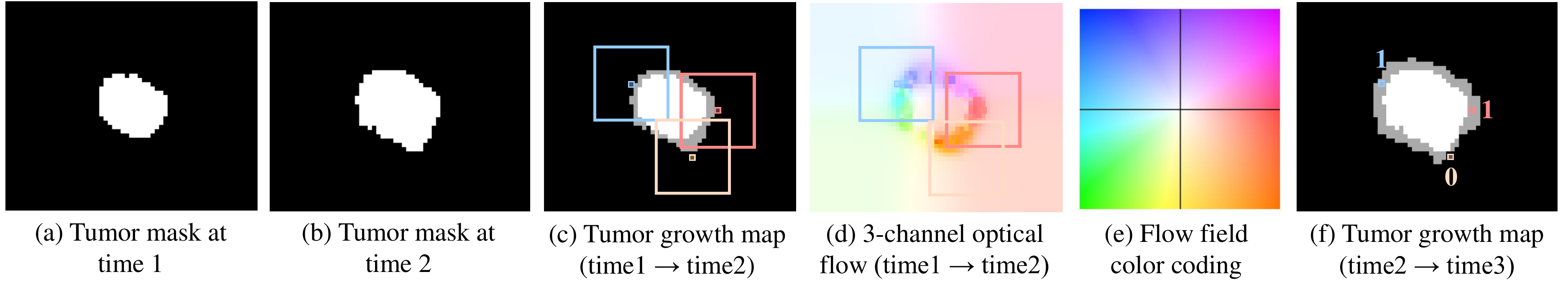}
   \end{center} \vspace{-3mm}
   \caption[example] 
   { \label{figOL} 
An example of color-coded optical flow image (d) generated based on the tumor mask pair at time 1 (a) and time 2 (b). The flow field color coding map is shown in (e), where hue indicates orientation and saturation indicates magnitude. In the tumor growth maps (c) and (f), white indicates the previous tumor region and gray indicates the newly grown tumor region. In (c) and (d), three non-tumor voxels and their surrounding image patches are highlighted by three colors, which indicate the colors of these voxels in (d). The blue and red voxels indicate left and right growing trend and both become tumors at time 3 (f), while the pink voxel indicates very small motion and is still non-tumor at time 3 (f). Also note that although some voxels show tiny motion (e.g., lower-left location) between time1 and time2, they grow faster from time2 to time3, indicating the nonlinear growth pattern of tumors.} 
   \end{figure*} 

More specifically, for a pair of consecutive tumor mask images at time1 and time2 (Fig. \ref{figOL} (a)-(b)), we use the algorithm in \cite{brox2004high} for optical flow estimation. The computed dense optical flow maps are a set of spatially coordinated displacement vector fields, which capture the displacement movements for all matched pairs of voxels from time1 to time2. 
By utilizing the color encoding scheme for flow visualization in \cite{baker2011database,liu2009beyond}, the magnitude and orientation of the vector field can be formed as a 3-channel color image (Fig. \ref{figOL} (d)). 
As depicted in the color coding map (Fig. \ref{figOL} (e)), the magnitude and orientation are represented by saturation and hue, respectively. This is a redundant but expressive visualization for explicitly capturing the motion dynamics of all corresponding voxels at different time points. Such a representation is also naturally fit for a ConvNet. The optical flow maps computed between raw CT image pairs may be noisy due to the inconsistent image appearance of tumors and surrounding tissues across two time points. Therefore, a binary tumor mask pair is used to estimate the optical flow due to it provides the growing trend of tumor mass. It should be mentioned that both the expansion and shrink motion can be coded in the 3-channel image.

However, such a representation of tumor growth motion has a potential limitation -- both the voxels locate around the tumor center and at background have very small motion, which may confuse the ConvNet. Therefore, we additionally provide the past (time1 and time2) locations of tumor by adding a tumor growth map (Fig. \ref{figOL} (c)) as the 4$^{th}$ input channel. Specifically, voxels belong to the overlap region of time1 and time2, newly growing (expansion) region, shrink region, and background are assigned values of 255, 170, 85, and 0, respectively. This strategy implicitly indicates the probabilities of voxels to be tumor or not in the future.

\subsubsection{Network Architecture}
\label{exparchitecture}
The expansion subnetwork has the same architecture as its invasion counterpart (cf. Section \ref{invarchitecture} and lower panel in Fig. \ref{figlatefusion}), and is trained to learn from our motion-based representations and infer the future involvement regions of the tumor. This network is trained from scratch on different time point configurations (\zzz{(}time1$\rightarrow$time2\zzz{)}/time3) of the population data set. In \cite{weizman2012prediction}, optical flow is used to predict the future tumor position in a scan, and the future motion of a voxel is directly predicted by a linear combination of its past motions, which may be over simplified. Our main difference is that the prediction is based on the nonlinear ConvNet learning of 2D motion and tumor growth maps where boundary/morphological information in a local region surrounding each voxel is maintained.

   \begin{figure*}[!t]
   \begin{center}
   \includegraphics[width=13.8cm]{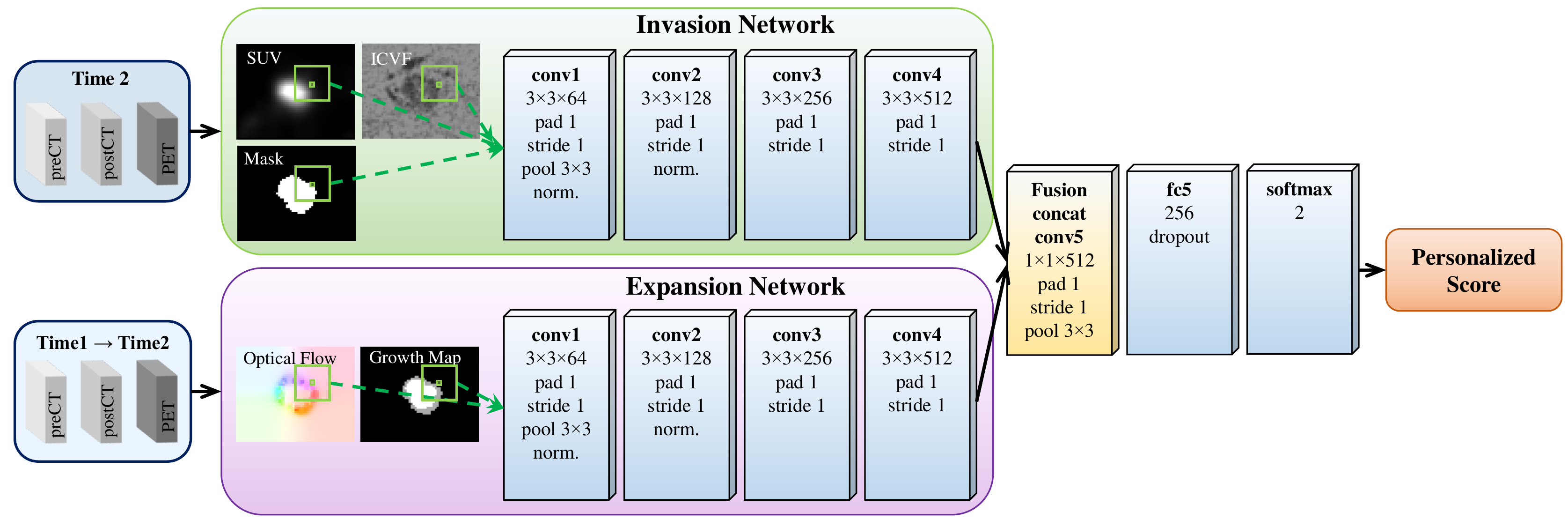}
   \end{center} \vspace{-3mm}
   \caption[example] 
   { \label{figend2end}  
Two-stream end-to-end fusion of the invasion and expansion networks for predicting tumor growth. The (convolution) fusion is after the $conv4$ (ReLU4) layer.
} \vspace{-3mm}
   \end{figure*} 

\subsection{Fusing Invasion and Expansion Networks}
To take advantage of the invasion-expansion information, we study a number of ways of fusing the invasion and expansion networks. Different fusion strategies result in significant different number of parameters in the networks.

\subsubsection{Two-Stream Late Fusion}
The two-stream architecture treats the appearance and motion cues separately and makes the prediction respectively. The fusion is achieved by averaging decision/softmax scores of two subnetworks, as shown in Fig. \ref{figlatefusion}. This method is denoted as \textit{late fusion}. The invasion and expansion subnetworks are trained on all time-point pairs (time1/time2 and time2/time3) and triplets (\zzz{(}time1$\rightarrow$time2\zzz{)}/time3) of the population data, respectively. Since they are trained independently, late fusion is not able to learn the voxel-wise correspondences between invasion and expansion features, i.e., registering appearance and motion cues. For example, what are the cell density and energy when a local voxel exhibits fast growing trend? Late fusion doubles the number of network parameters compared to invasion or expansion subnetworks only.
 
\subsubsection{One-Stream Early Fusion}
In contrast to late fusion, we present an \textit{early fusion} architecture, which directly stacking the 3-channel invasion and 4-channel expansion images as a 7-channel input to the ConvNet. The same network architecture as invasion/expansion network is used. Different from late fusion, early fusion can only be trained on time2/time3 pairs (without time1/time2 pairs) along with triplets (\zzz{(}time1$\rightarrow$time2\zzz{)}/time3) of the population data. Therefore, less training samples can be used. Early fusion is able to establish voxel-wise correspondences. However, it leaves the correspondence to be defined by subsequent layers through learning. As a result, information in the motion image may not be able to be well captured by the network, since there is more variability in the appearance images (i.e., SUV and ICVF). Early fusion keeps almost the same number of parameters as a single invasion or expansion network.

\subsubsection{Two-Stream End-to-End Fusion}
To jointly learn the nonlinear static and dynamic tumor information while allocating enough network capacity to both appearance and motion cues, we introduce a two-stream \textit{end-to-end} fusion architecture. As shown in Fig. \ref{figend2end}, the two subnetworks are connected by a fusion layer that adds a convolution on top of their $conv4$ layers. More specifically, the fusion layer first concatenates the two feature maps generated by $conv4$ (after ReLU4) and convolves the stacked data with $1\times 1\times 512$ convolution filters with padding and stride of 1, then ReLU5 is attached and max-pooling $3\times 3$ is performed. The outputs of the fusion layer are fed into a subsequent fully-connected layer ($fc5$). As such, the fusion layer is able to learn correspondences of two compact feature maps that minimize a joint loss function. Fusion at ReLU4 instead of $fc$ layer is because the spatial correspondences between invasion and expansion are already collapsed at the $fc$ layer; fusion at the last $conv$ layer has been demonstrated to have higher accuracy in compared to at earlier $conv$ layers \cite{feichtenhofer2016convolutional}. End-to-end framework is trained on the same time pairs and triplets as early fusion, without time1/time2 pairs compared to late fusion. End-to-end fusion removes nearly half of the parameters in the late fusion architecture as only one tower of $fc$ layer is used after fusion.

\subsection{Personalizing Invasion and Expansion Networks}
Predictive model personalization is a key step of model-based tumor growth prediction \cite{chen2013kidney,liu2014patient,wong2015tumor,wong2017pancreatic}. In statistical learning, model validation is a natural way to optimize the pre-trained model. Particularly, given tumor status at time1 and time2 already known (predict time3), the model personalization includes two steps. In the first step, the invasion network is trained on population data and time1/time2 of the target patient is used as validation. Training is terminated after a pre-determined number (30) of epochs, after which the model snapshot with the lowest validation loss on the target patient data is selected. Since there are no corresponding validation datasets for the expansion network, early fusion, and end-to-end fusion, their trainings are terminated after the empirical number of 20 epochs, in order to reduce the risk of overfitting. 

To better personalize the invasion network to the target patient, we propose a second step that optimizes an objective function which measures the agreement between any predicted tumor volume and its corresponding future ground truth volume on the target patient. This is achieved by directly applying the invasion network to voxels in a tumor growth zone in the personalization volume, and later thresholding the probability values of classification outputs to reach the best objective function. Dice coefficient measures the agreement between ground truth and predicted volumes, and is used as the objective function is this study:
\begin{equation}\label{dice}
{\rm Dice} = \frac{2 \times \rm TPV}{V_{pred} + V_{gt}}
\end{equation}
where TPV is the true positive volume -- the overlapping volume between the predicted tumor volume $V_{pred}$ and the ground truth tumor volume $V_{gt}$. The tumor growth zone is set as a bounding box surrounding the tumor, with pixel distances $N_{x}$, $N_{y}$, and $N_{z}$ from the tumor surface in the $x$, $y$, and $z$ directions, respectively. The personalized threshold of invasion network is also used for expansion network and the three fusion networks.

\subsection{Predicting with Invasion and Expansion Networks}
During testing, given the imaging data at time1 and time2 for the target patient, one of the frameworks, the personalized invasion network, expansion network, late fusion, early fusion, or end-to-end fusion could be applied to predict the scores for every voxels in the growth zone at the future time3. The static information from time2 serves as invasion information, while the motion/change information between time1 and time2 represents the expansion information. Late fusion and end-to-end fusion feed the static and motion information to invasion and expansion subnetworks, separately, while early fusion concatenates both static and motion information as input to a one-stream ConvNet.

\section{Experimental Methods}

\subsection{Data and Protocol}
Ten patients (six males and four females) with von Hippel-Lindau (VHL) disease, each with a pancreatic neuroendocrine tumor (PanNET), are studied in this paper. The VHL-associated PanNETs are commonly found to be nonfunctioning with malignant (cancer) potential \cite{keutgen2016evaluation}, and can often be recognized as well-demarcated and solid masses through imaging screens \cite{wolfgang2013recent}. For the natural history of this kind of tumor, around 60\% patients demonstrate nonlinear tumor growth, 20\% stable and 20\% decreasing (over a median follow-up duration of 4 years) \cite{weisbrod2014assessment}. Treatments of PanNETs include active surveillance, surgical intervention, and medical treatment. Active surveillance is undertaken if a PanNET does not reach 3 cm in diameter or a tumor-doubling time $<$500 days; other wise the PanNET should be resected due to high risk of metastatic disease \cite{keutgen2016evaluation}. Medical treatment (e.g., everolimus) is for the intermediate-grade (PanNETs with radiologic documents of progression within the previous 12 months), advanced or metastatic disease \cite{yao2011everolimus}. Therefore, patient-specific prediction of spatial-temporal progression of PanNETs at earlier stage is desirable, as it will assist making decision within different treatment strategies to better manage the treatment or surgical planning. 

In our dataset, each patients has three time points of contrast-enhanced CT and FDG-PET imaging spanning three to four years, with the time interval of $405 \pm 133$ days (average $\pm$ std.). The average age of the patients at time1 is 46.9 $\pm$ 13.2 years. The image pixel sizes range between $0.68 \times 0.68 \times 1$ mm$^3$ ---  $0.98 \times 0.98 \times 1$ mm$^3$ for CT and $2.65 \times 2.65 \times 1.5$ mm$^3$ --- $4.25 \times 4.25 \times 3.27$ mm$^3$ for PET. The tumor growth information of all patients is shown in Table \ref{tabletumorinfor}. Most tumors are slow growing, while two are more aggressive and two experience shrinkage. Some tumors keep a similar growing rate as their past trend, while others have varying growing rates.

\begin{table}[t]
\tabcolsep=0.11cm
\caption{Tumor information at the 1st, 2nd, and 3rd time points of ten patients.}
\label{tabletumorinfor}
\footnotesize
\begin{tabular}{ p{1.5cm}p{0.7cm}p{1.4cm}p{0.7cm}p{1.4cm}p{1.8cm} }
    \toprule
    & \multicolumn{2}{l}{1st-2nd} & \multicolumn{2}{l}{2nd-3rd} &   \\
    \cmidrule(lr){2-3} \cmidrule(lr){4-5}
    {Patient ID} & {Days} & {Growth (\%)}  & {Days} & {Growth (\%)} & {Size (cm$^{3}$, 3rd)}  \\
    \midrule
    1 & 384 & 34.6 & 804 & 33.4 & 2.3 \\
    2 & 363 & 15.3 & 363 & 10.7 & 1.4 \\
    3 & 378 & 18.9 & 372 & 7.5 & 0.4 \\
		4 & 364 & 150.1 & 364 & 28.9 & 3.1 \\
    5 & 426 & 41.5 & 420 & 68.6 & 3.8 \\
    6 & 372 & 7.4 & 360 & 12.5 & 6.3 \\
		7 & 384 & 13.6 & 378 & -3.9 & 1.6 \\
    8 & 168 & 18.7 & 552 & 18.7 & 3.2 \\
    9 & 363 & 16.9 & 525 & 34.7 & 0.3 \\
		10 & 196 & -28.9 & 567 & 17.7 & 0.9 \\
    \bottomrule
\end{tabular}

\end{table}

\subsection{Implementation Details}
A total of  45,989 positive and 52,996 negative image patches is used for the invasion network in late fusion, and 23,448 positive and 25,896 negative image patches for both the invasion network and expansion network in other fusion (i.e., early and end-to-end), extracted from 10 patients. Each image patch is subtracted by the mean image patch over the training set. Data augmentation is not performed since we could not observe improvements in a pilot study. The following hyperparamaters are used: initial learning rate -- 0.001, decreased by a factor of 10 at every tenth epoch; weight decay -- 0.0005; momentum -- 0.9; mini-batch size -- 512. We use an aggressive dropout ratio of 0.9 to improve generalization. Lower dropout ratios (e.g., 0.5) do not decrease performance significantly. The ConvNets are implemented using Caffe platform \cite{jia2013caffe}. The parameters for tumor growth zone are set as $N_{x}=3$, $N_{y}=3$, and $N_{z}=3$ for prediction speed concern. We observe that the prediction accuracy is not sensitive to the choice of these parameters, e.g., $N_{x|y|z}\geq4$ results in similar performance. For the model personalization via Dice coefficient objective function, we vary the model thresholding values in the range of [0.05, 0.95] with 0.05 intervals. The proposed method is tested on a DELL TOWER 7910 workstation with 2.40 GHz Xeon E5-2620 v3 CPU, 32 GB RAM, and a Nvidia TITAN X Pascal GPU of 12 GB of memory.

\subsection{Evaluation Methods}
\label{evaluation}
The proposed method is evaluated using leave-one-out cross-validation. In each of the 10 evaluations, 9 patients are used as the population training data to learn the population trend, the time1/time2 of the remaining patient is used as the personalization data set for invasion network, and time3 of the remaining patient as the to-be-predicted testing set. We obtain the model's final performance values by averaging results from the 10 cross validations. The numbers of parameters in each of the proposed network are reported, and the prediction performances are evaluated using measurements at the third time point by recall, precision, Dice coefficient (defined in Eq. \ref{dice}), and RVD (relative volume difference) as in \cite{wong2017pancreatic,zhang2017personalized}.
\begin{equation}
{\rm recall} = \frac{TPV}{V_{gt}};~~~{\rm precision} = \frac{TPV}{V_{pred}};~~~{\rm RVD} = \frac{V_{pred} - V_{gt}}{V_{gt}}
\end{equation}

To establish a benchmark for comparisons, we implement a \textit{linear growth} model that assumes that tumors would keep their past growing trend in the future. More specifically, we first compute the radial expansion/shrink distances on tumor boundaries between the first and second time points, and then expand/shrink the tumor boundary at the second time point to predict the third with the same radial distances. Furthermore, we compare the accuracy and efficiency of our method with two state-of-the-art tumor growth prediction methods \cite{wong2017pancreatic,zhang2017personalized} which have been evaluated on a subset (7 patients, without patient 4, 7, 10 in Table \ref{tabletumorinfor}) of the same dataset. Finally, to show the importance of model personalization, the prediction performance with and without our personalization method (i.e., optimizing Eq. (\ref{dice})) are compared.

\section{Results}

\begin{figure}[!t]
   \begin{center}
     \includegraphics[width=8.5cm]{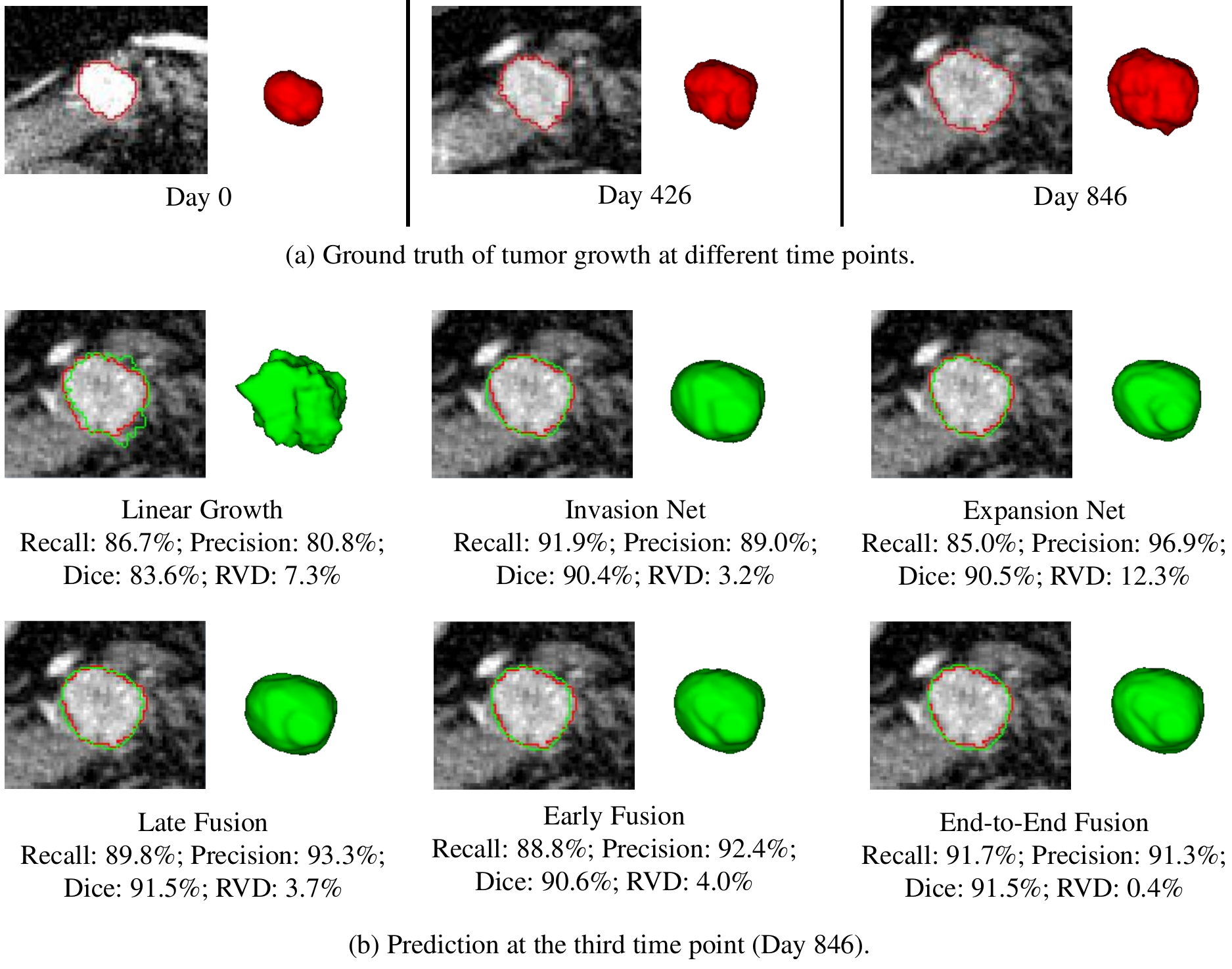}
      \end{center} 
   \caption{ 
An example (patient 5) shows the tumor growth prediction by our individual and fusion networks. (a) The segmented (ground truth) tumor contours and volumes at different time points. (b) The prediction results at the third time point, with red and green represent ground truth and predicted tumor boundaries, respectively.
} \label{figresultp5} 
   \end{figure}

\begin{figure}[!t]
   \begin{center}
     \includegraphics[width=8.5cm]{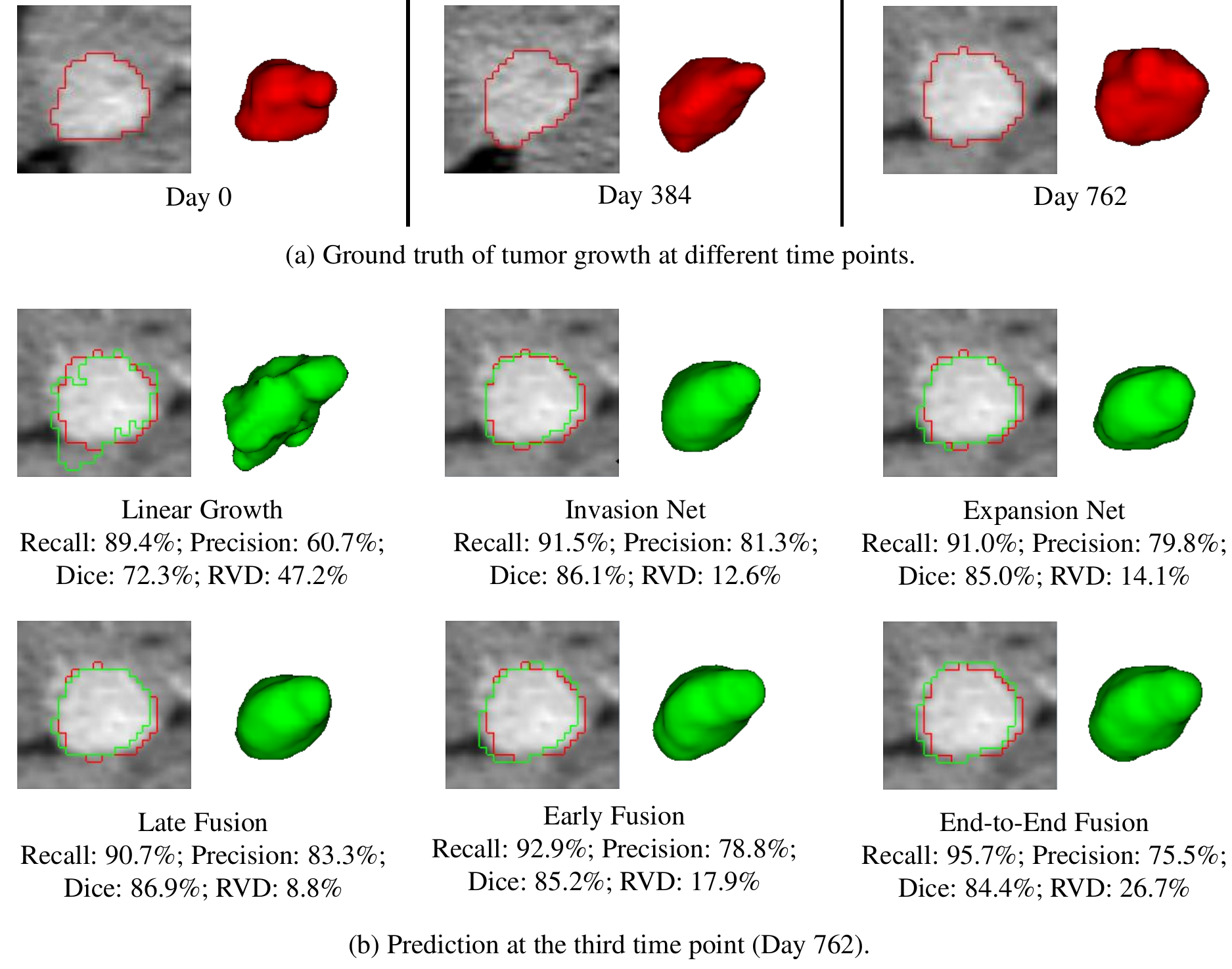}
      \end{center} 
   \caption{ 
An example (patient 7) shows the tumor growth prediction by our individual and fusion networks. (a) The segmented (ground truth) tumor contours and volumes at different time points. (b) The prediction results at the third time point, with red and green represent ground truth and predicted tumor boundaries, respectively.
} \label{figresultp7} 
   \end{figure}

Fig. \ref{figresultp5} shows a good prediction results obtained by our individual and fusion networks. In this example (patient 5), the tumor is growing in a relatively steady trend. Therefore, all the predictive models including the linear model can achieve promising prediction accuracy. Our methods, especially the network fusions (e.g., late and end-to-end) balance the recall and precision of individual networks, yield the highest accuracy. Fig. \ref{figresultp7} shows the results of patient 7. In this case, the tumor demonstrates a nonlinear growth trend, and its size first increases from time1 to time2 but decreases a little bit from time2 to time3. Therefore, all the personalized predictive models overpredicted the tumor size (recall is higher than precision). However, our models especially the two-stream late fusion can still generate promising prediction result. 

Table \ref{performance10} presents the overall prediction performance on 10 patients. Compared to the baseline linear growth method, all our methods show substantially higher performance. The performance of invasion and expansion networks are comparable. Fusion of the two networks can further improve the prediction accuracy, especially for the RVD measure. Two-stream late fusion achieves the highest mean values with Dice coefficient of $85.9 \pm 5.6\%$ and RVD of $8.1 \pm 8.3\%$, but requires nearly twice of the model parameters in compared to \zzz{early} fusion. \zzz{End-to-end fusion has the second highest accuracy with much less network parameters than late fusion.} Nevertheless, this suggests that the mechanism of fusion ConvNets leverages the complementary relationship between static and dynamic tumor information.

\begin{table*}[!t] 
\scriptsize
\centering
\caption{Overall performance on 10 patients -- baseline linear predictive model, invasion network, expansion network, early fusion, late fusion, and end-to-end fusion. Results are estimated by the recall, precision, Dice coefficient, and relative volume difference (RVD), and are reported as: mean $\pm$ std [min, max]. The numbers of parameters for each model are provided.
}
\label{performance10}
\begin{tabular}{p{1.4cm}p{2.5cm}p{2.5cm}p{2.5cm}p{2.5cm}p{1.3cm}}
\hline
 & Recall (\%) & Precision (\%) & Dice (\%) & RVD (\%) & \#parameters \\
\hline
Linear & 84.5$\pm$\textbf{7.0} [\textbf{73.3}, \textbf{97.3}] & 69.5$\pm$8.0 [60.7, 82.3] & 75.9$\pm$5.4 [69.5, 85.0] & 23.1$\pm$18.5 [5.9, 58.8] & - \\

Invasion & 86.9$\pm$9.4 [63.7, 97.0] & 83.3$\pm$5.6 [74.7, 90.2] & 84.6$\pm$5.1 [73.0, 90.4] & 11.5$\pm$11.3 [2.3, 30.0] & 8.11M \\

Expansion & \textbf{87.6}$\pm$8.6 [68.3, 96.5] & 82.9$\pm$7.6 [76.5, \textbf{97.2}] & 84.8$\pm$5.4 [73.2, 91.1] & 13.8$\pm$\textbf{6.3} [1.0, 23.5] & 8.11M\\

Early fusion & 86.4$\pm$7.9 [66.6, 94.8] & 84.7$\pm$5.8 [77.0, 92.7] & 85.2$\pm$5.2 [73.9, 90.6] & 9.2$\pm$7.3 [2.4, \textbf{19.6}] & 8.11M \\

Late fusion & 86.9$\pm$8.8 [64.0, 95.5] & \textbf{85.5}$\pm$\textbf{4.9} [\textbf{78.6}, 91.3] & \textbf{85.9}$\pm$5.6 [72.8, \textbf{91.7}] & \textbf{8.1}$\pm$8.3 [1.0, 24.2] & 16.22M \\

End-to-end & 87.5$\pm$8.1 [70.0, 96.9] & 84.1$\pm$5.6 [75.5, 91.3] & 85.5$\pm$\textbf{4.8} [\textbf{76.5}, 91.5] & 9.0$\pm$10.1 [\textbf{0.3}, 26.7] & 10.18M \\
\hline
\end{tabular}
\end{table*}

Table \ref{performance7} compares our methods with two state-of-the-art methods \cite{wong2017pancreatic,zhang2017personalized} on a subset (seven patients) of our data. Out of ten patients, three patients (patient 4, 7, and 10 in Table \ref{tabletumorinfor}) with aggressive and shrink tumors are not included in the experiment. As a result, the performances on seven patients (Table \ref{performance7}) are better than that on ten patients (Table \ref{performance10}). Our single network can already achieve better accuracy than the model-based method (i.e., EG-IM) \cite{wong2017pancreatic}, especially the invasion network has a much lower/better RVD than \cite{wong2017pancreatic}. This demonstrates the highly effectiveness of ConvNets (learning invasion information) in future tumor volume estimation. Network fusions further improve the accuracy and achieve comparable performance with the group learning method \cite{zhang2017personalized}, which benefits results from integrating the deep features, hand-crafted features, and clinical factors into a SVM based learning framework. Again, the two-stream late fusion performs the best among the proposed three fusion architectures, with Dice coefficient of $86.8 \pm 3.4\%$ and RVD of $6.6 \pm 7.1\%$.

\begin{table*}[!t] 
\scriptsize
\centering
\caption{Comparison of performance on 7 patients -- baseline linear predictive model, state-of-the-art model-based \cite{wong2017pancreatic}, statistical group learning \cite{zhang2017personalized}, and our models. Results are estimated by the recall, precision, Dice coefficient, and relative volume difference (RVD), and are reported as: mean $\pm$ std [min, max]. EG-IM-FEM* has higher performance than EG-IM, but it has some issues mentioned by the authors (Sec. VI in \cite{wong2017pancreatic}).
}
\label{performance7}
\begin{tabular}{p{2.3cm}p{2.5cm}p{2.5cm}p{2.5cm}p{2.5cm}}
\hline
 & Recall (\%) & Precision (\%) & Dice (\%) & RVD (\%)\\
\hline
Linear & 84.3$\pm$\textbf{3.4} [78.4, 88.2] & 72.6$\pm$7.7 [64.3, 82.1] & 77.3$\pm$5.9 [72.3, 85.1] & 16.7$\pm$10.8 [5.2, 34.3] \\

EG-IM \cite{wong2017pancreatic} & 83.2$\pm$8.8 [69.4, 91.1] & \textbf{86.9}$\pm$8.3 [74.0, \textbf{97.8}] & 84.4$\pm$4.0 [79.5, \textbf{92.0}] & 13.9$\pm$9.8 [3.6, 25.2] \\

EG-IM-FEM* \cite{wong2017pancreatic} & 86.8$\pm$5.8 [77.6, 96.1] & 86.3$\pm$8.2 [72.7, 96.5] & 86.1$\pm$\textbf{3.2} [\textbf{82.8}, 91.7] & 10.8$\pm$11.2 [2.3, 32.3] \\

Group learning \cite{zhang2017personalized}  & 87.9$\pm$5.0 [81.4, 94.4] & 86.0$\pm$5.8 [\textbf{78.7}, 94.5] & \textbf{86.8}$\pm$3.6 [81.8, 91.3] & 7.9$\pm$\textbf{5.4} [2.5, 19.3] \\

Invasion & 88.1$\pm$4.6 [81.4, 94.3] & 84.4$\pm$5.6 [75.0, 90.2] & 86.1$\pm$3.6 [80.8, 90.4] & 6.6$\pm$8.5 [2.3, 25.8] \\

Expansion & \textbf{90.1}$\pm$6.3 [79.1, 96.5] & 81.9$\pm$6.9 [76.5, 96.9] & 85.5$\pm$3.8 [78.7, 90.5] & 14.2$\pm$7.6 [1.0, 23.5] \\

Early fusion & 88.2$\pm$4.2 [81.9, 94.8] & 85.2$\pm$6.5 [77.0, 92.7] & 86.5$\pm$4.0 [80.7, 90.6] & 7.5$\pm$6.1 [2.5, \textbf{19.0}] \\

Late fusion & 89.1$\pm$4.3 [\textbf{83.4}, 95.5] & 84.9$\pm$\textbf{5.2} [78.6, 93.3] & \textbf{86.8}$\pm$3.4 [81.8, 91.5] & \textbf{6.6}$\pm$7.1 [1.0, 21.5] \\

End-to-end & 88.8$\pm$5.9 [79.1, \textbf{96.9}] & 84.8$\pm$5.6 [77.8, 91.3] & 86.6$\pm$4.4 [80.5, 91.5] & \textbf{6.6}$\pm$8.3 [\textbf{0.4}, 24.4] \\
\hline
\end{tabular}
\end{table*}

The proposed two-stream late fusion ConvNets (our other architectures are even faster) requires $\sim5$ mins for training and personalization, and $15$ s for prediction per patient, on average -- significantly faster than the model-based approach in \cite{wong2017pancreatic} ($\sim 24$ hrs -- model personalization; 21 s -- simulation), and group learning method in \cite{zhang2017personalized} ($\sim 3.5$ hrs -- model training and personalization; 4.8 mins -- prediction).  

The comparison between with and without personalization is shown in Table \ref{personalization}. The personalization process significantly improves the prediction performance especially for predicting the future tumor volume (i.e., RVD), demonstrating its crucial role in tumor growth prediction.

\begin{table}[!t] 
\scriptsize
\centering
\caption{Comparison between with and without personalization (w/o P) on all 10 patients. For more concisely, only Dice coefficient, and relative volume difference (RVD) are reported.
}
\label{personalization}
\begin{tabular}{p{2.5cm}p{2.5cm}p{2.5cm}}
\hline
 & Dice (\%) & RVD (\%)\\
\hline
Invasion w/o p & 77.5$\pm$7.8 [65.1, 90.5] & 51.4$\pm$31.8 [3.0, 105.3] \\

Expansion w/o p & 78.0$\pm$7.4 [67.7, 90.0] & 50.6$\pm$24.1 [15.6, 93.2] \\

Early fusion w/o p & 81.3$\pm$6.4 [70.7, 90.2] & 36.3$\pm$24.6 [10.1, 81.7] \\

Late fusion w/o p & 78.3$\pm$6.9 [67.7, 90.3] & 49.9$\pm$25.9 [11.5, 92.7] \\

End-to-end w/o p & 80.8$\pm$6.4 [72.8, 91.1] & 38.5$\pm$24.2 [5.3, 74.6] \\
\hline

Invasion & 84.6$\pm$5.1 [73.0, 90.4] & 11.5$\pm$11.3 [2.3, 30.0] \\

Expansion & 84.8$\pm$5.4 [73.2, 91.1] & 13.8$\pm$6.3 [1.0, 23.5] \\

Early fusion & 85.2$\pm$5.2 [73.9, 90.6] & 9.2$\pm$7.3 [2.4, 19.6] \\

Late fusion & 85.9$\pm$5.6 [72.8, 91.7] & 8.1$\pm$8.3 [1.0, 24.2] \\

End-to-end & 85.5$\pm$4.8 [76.5, 91.5] & 9.0$\pm$10.1 [0.3, 26.7] \\
\hline
\end{tabular}
\end{table}

\section{Discussions}

Tumor growth prediction is a biophysics process and has long been solved via mathematical modeling. In this paper, we tackle this task using novel ConvNet architectures of convolutional invasion and expansion neural networks, with respect to the cell invasion and mass-effect processes, jointly. Our new approach demonstrates promising accuracy and highly efficiency. Although the small data size does not permit statistical testing, our \zzz{prediction} method clearly shows higher mean and lower std. than the state-of-the-art modeling method \cite{wong2017pancreatic} \zzz{when the same preprocessing (e.g., registration, segmentation) procedure as the pipeline in \cite{wong2017pancreatic} is used}. 

Besides using deep learning instead of mathematical modeling, the main difference against \cite{wong2017pancreatic} is using the information from other patients as population prior learning followed by personalization using the same patient's 1st and 2nd time point data. Ref. \cite{wong2017pancreatic} does not use other patients’ information but directly trains on the 1st and 2nd time points to simulate the 3rd time point tumor growth for the same patient, which, in some sense, may be more likely to overfit. Our prior population learning may behave as a beneficial regularization to constrain the personalized predictive model. As such, compared to the model-based prediction, our method is better at predicting the tumors which will have a different (even opposite) growing trend to their past trend. An example can be seen in Fig. \ref{figresultp7} for patient 7. Actually, on another challenging case -- patient 4 (aggressive growth from time1 to time2), our late fusion yields very promising prediction with Dice of 91.7\% and RVD of 2.2\%. The worst case is for patient 10 (shrink from time1 to time2), late fusion has a Dice of 72.8\% and RVD of 24.2\%. \zzz{We also investigate the performance of our method without the population data. For example, we only train and personalize the invasion network on a patient's target data (time1/time2 pair) using the same strategy proposed in section \ref{convIEnets}, and then predict tumor growth at time3. The overall Dice and RVD on 10 patients are 74.7\% and 32.2\%, respectively, substantially worse than the invasion network with population learning (Dice = 84.6\%, RVD = 11.5\%).}

Model personalization is one of the main novelties of our deep learning based method. This strategy ensures a robust prediction performance, and may subsequently benefit more from the following directions. 1) The Dice coefficient is used as the objective function. Using RVD as the objective function (as in \cite{zhang2017personalized}) actually result in comparable but (maybe) slightly lower performance. For example, for the two-stream late fusion, using RVD as objective function of personalization will result in 0.1\% lower of Dice and 1.9\% larger of RVD metrics in prediction. The prediction performances with different objective functions (e.g., weighted combination of Dice and RVD) need further investigation. 2) Since there is no validation data for the expansion network (also for other fusion networks), its personalization empirically follows the invasion network. A better personalization could be achieved if more time points would be available (e.g., tumor status at time1, 2, and 3 already known, predict time4). This is actually a common scenario in practice for many kinds of tumors, such as predicting time7 based on time1-6 for kidney tumors \cite{chen2013kidney}, and predicting time5 given time1-4 known for brain tumors \cite{weizman2012prediction}. Therefore, we could expect better performance given a dataset spanning more time points. 3) Our predictive models perform much worse if without personalization. Besides the importance of personalization, this may be caused by the patch sampling strategy, which proportionally under-samples negative samples in a predefined bounding box (section \ref{invasionpatch}). As a result, some `easy-negatives' (far from tumor boundary) are involved in the training set, lowering the ConvNet's capacity in discriminating some `hard-negatives' (close to tumor boundary). Restricting the patch sampling to the range close to the tumor boundary without under-sampling has a potential to improve this issue. 

A simple linear prediction approach shows the worst performance among all the models. This is in agreement with the fact that PanNETs demonstrate nonlinear growth \cite{keutgen2016evaluation,weisbrod2014assessment}. The two-stream late fusion performs slightly better than the one-stream early fusion and two-stream end-to-end fusion architectures (Table \ref{performance10} and Table \ref{performance7}). Probably, the reason is that the late fusion is trained on more training samples, in compared to early and end-to-end fusion which cannot use samples of time1/time2 pairs for model training.

A potential limitation of the current method is that the crucial tumor biomechanical properties, such as tissue biomechanical strain, is not considered. These limitations could be addressed by fusing our proposed deep learning method with traditional biomechanical model-based methods. Finally, although our dataset (ten patients) is already the largest for this kind of research, it is still too small. Therefore, some of the results and discussions should be treated with caution. \zzz{To evaluate our method, we conduce a leave-one-patient-out cross-validation, which is a popular error estimation procedure when the sample size is small. Furthermore, our method has a personalization stage where patient specific data is employed to optimize the model generated by the training data. This strategy can somehow alleviate the small training set problem. Nevertheless, more training data will likely enhance our convolutional invasion and expansion networks (an end-to-end deep learning model). As an ongoing clinical trial in NIH, we are collecting more longitudinal panNET data and kidney tumor data. We will validate and extend our method on the new data.}

\section{Conclusions}
In this paper, we show that deep ConvNets can effectively represent and learn both cell invasion and mass-effect in tumor growth prediction. Composite images encoding static and dynamic tumor information are fed into our ConvNet architectures to predict the future involvement region of pancreatic tumors. Our method surpasses the state-of-the-art mathematical model-based method \cite{wong2017pancreatic} in both speed and accuracy, and is much more efficient than our recently proposed group learning method \cite{zhang2017personalized}. The invasion and expansion networks alone predict the tumor growth at higher accuracies than \cite{wong2017pancreatic}, and our proposed fusion architectures further improve the prediction accuracy. \zzz{Two-stream end-to-end fusion might be a trade-off between accuracy and generalization compared with early and late fusions.}

\section*{Acknowledgment}
The authors gratefully thank Nvidia for the TITAN X Pascal GPU donation. They would also like to thank Ms. Isabella Nogues for proofreading this article.

\ifCLASSOPTIONcaptionsoff
  \newpage
\fi

{\small\small
\bibliographystyle{IEEEtran}
\bibliography{LingRef}

\begin{thebibliography}{10}
\providecommand{\url}[1]{#1}
\csname url@samestyle\endcsname
\providecommand{\newblock}{\relax}
\providecommand{\bibinfo}[2]{#2}
\providecommand{\BIBentrySTDinterwordspacing}{\spaceskip=0pt\relax}
\providecommand{\BIBentryALTinterwordstretchfactor}{4}
\providecommand{\BIBentryALTinterwordspacing}{\spaceskip=\fontdimen2\font plus
\BIBentryALTinterwordstretchfactor\fontdimen3\font minus
  \fontdimen4\font\relax}
\providecommand{\BIBforeignlanguage}[2]{{%
\expandafter\ifx\csname l@#1\endcsname\relax
\typeout{** WARNING: IEEEtran.bst: No hyphenation pattern has been}%
\typeout{** loaded for the language `#1'. Using the pattern for}%
\typeout{** the default language instead.}%
\else
\language=\csname l@#1\endcsname
\fi
#2}}
\providecommand{\BIBdecl}{\relax}
\BIBdecl

\bibitem{warburg1956origin}
O.~Warburg, ``On the origin of cancer,'' \emph{Science}, vol. 123, no. 3191,
  pp. 309--314, 1956.

\bibitem{friedl2012classifying}
P.~Friedl, J.~Locker, E.~Sahai, and J.~E. Segall, ``Classifying collective
  cancer cell invasion,'' \emph{Nature Cell Biology}, vol.~14, no.~8, pp.
  777--783, 2012.

\bibitem{keutgen2016evaluation}
X.~M. Keutgen, P.~Hammel, P.~L. Choyke, S.~K. Libutti, E.~Jonasch, and
  E.~Kebebew, ``Evaluation and management of pancreatic lesions in patients
  with von hippel-lindau disease,'' \emph{Nature Reviews Clinical Oncology},
  vol.~13, no.~9, pp. 537--549, 2016.

\bibitem{liu2014patient}
Y.~Liu, S.~Sadowski, A.~Weisbrod, E.~Kebebew, R.~Summers, and J.~Yao, ``Patient
  specific tumor growth prediction using multimodal images,'' \emph{Medical
  Image Analysis}, vol.~18, no.~3, pp. 555--566, 2014.

\bibitem{wong2015tumor}
K.~C. Wong, R.~M. Summers, E.~Kebebew, and J.~Yao, ``Tumor growth prediction
  with reaction-diffusion and hyperelastic biomechanical model by physiological
  data fusion,'' \emph{Medical Image Analysis}, vol.~25, no.~1, pp. 72--85,
  2015.

\bibitem{wong2017pancreatic}
K.~C.~L. Wong, R.~M. Summers, E.~Kebebew, and J.~Yao, ``Pancreatic tumor growth
  prediction with elastic-growth decomposition, image-derived motion, and
  {FDM}-{FEM} coupling,'' \emph{IEEE Transactions on Medical Imaging}, vol.~36,
  no.~1, pp. 111--123, 2017.

\bibitem{weisbrod2014assessment}
A.~B. Weisbrod, M.~Kitano, F.~Thomas, D.~Williams, N.~Gulati, K.~Gesuwan,
  Y.~Liu, D.~Venzon, I.~Turkbey, P.~Choyke \emph{et~al.}, ``Assessment of tumor
  growth in pancreatic neuroendocrine tumors in von hippel lindau syndrome,''
  \emph{Journal of the American College of Surgeons}, vol. 218, no.~2, pp.
  163--169, 2014.

\bibitem{swanson2000quantitative}
K.~R. Swanson, E.~Alvord, and J.~Murray, ``A quantitative model for
  differential motility of gliomas in grey and white matter,'' \emph{Cell
  Proliferation}, vol.~33, no.~5, pp. 317--329, 2000.

\bibitem{clatz2005realistic}
O.~Clatz, M.~Sermesant, P.-Y. Bondiau, H.~Delingette, S.~K. Warfield,
  G.~Malandain, and N.~Ayache, ``Realistic simulation of the {3D} growth of
  brain tumors in {MR} images coupling diffusion with biomechanical
  deformation,'' \emph{IEEE Transactions on Medical Imaging}, vol.~24, no.~10,
  pp. 1334--1346, 2005.

\bibitem{hogea2007modeling}
C.~Hogea, C.~Davatzikos, and G.~Biros, ``Modeling glioma growth and mass effect
  in 3{D} {MR} images of the brain,'' in \emph{MICCAI}.\hskip 1em plus 0.5em
  minus 0.4em\relax Springer, 2007, pp. 642--650.

\bibitem{hogea2008image}
------, ``An image-driven parameter estimation problem for a
  reaction--diffusion glioma growth model with mass effects,'' \emph{Journal of
  Mathematical Biology}, vol.~56, no.~6, pp. 793--825, 2008.

\bibitem{menze2011generative}
B.~H. Menze, K.~Van~Leemput, A.~Honkela, E.~Konukoglu, M.-A. Weber, N.~Ayache,
  and P.~Golland, ``A generative approach for image-based modeling of tumor
  growth,'' in \emph{IPMI}.\hskip 1em plus 0.5em minus 0.4em\relax Springer,
  2011, pp. 735--747.

\bibitem{chen2013kidney}
X.~Chen, R.~M. Summers, and J.~Yao, ``Kidney tumor growth prediction by
  coupling reaction--diffusion and biomechanical model,'' \emph{IEEE
  Transactions on Biomedical Engineering}, vol.~60, no.~1, pp. 169--173, 2013.

\bibitem{weizman2012prediction}
L.~Weizman, L.~Ben-Sira, L.~Joskowicz, O.~Aizenstein, B.~Shofty,
  S.~Constantini, and D.~Ben-Bashat, ``Prediction of brain {MR} scans in
  longitudinal tumor follow-up studies,'' in \emph{MICCAI}.\hskip 1em plus
  0.5em minus 0.4em\relax Springer, 2012, pp. 179--187.

\bibitem{morris2006learning}
M.~Morris, R.~Greiner, J.~Sander, A.~Murtha, and M.~Schmidt, ``Learning a
  classification-based glioma growth model using {MRI} data,'' \emph{Journal of
  Computers}, vol.~1, no.~7, pp. 21--31, 2006.

\bibitem{lecun2015deep}
Y.~LeCun, Y.~Bengio, and G.~Hinton, ``Deep learning,'' \emph{Nature}, vol. 521,
  no. 7553, pp. 436--444, 2015.

\bibitem{lecun1989backpropagation}
Y.~LeCun, B.~Boser, J.~S. Denker, D.~Henderson, R.~E. Howard, W.~Hubbard, and
  L.~D. Jackel, ``Backpropagation applied to handwritten zip code
  recognition,'' \emph{Neural Computation}, vol.~1, no.~4, pp. 541--551, 1989.

\bibitem{krizhevsky2012imagenet}
A.~Krizhevsky, I.~Sutskever, and G.~E. Hinton, ``Image{N}et classification with
  deep convolutional neural networks,'' in \emph{NIPS}, 2012, pp. 1097--1105.

\bibitem{greenspan2016guest}
H.~Greenspan, B.~van Ginneken, and R.~M. Summers, ``Guest editorial deep
  learning in medical imaging: Overview and future promise of an exciting new
  technique,'' \emph{IEEE Transactions on Medical Imaging}, vol.~35, no.~5, pp.
  1153--1159, 2016.

\bibitem{holger2016improving}
H.~R. Roth, L.~Lu, J.~Liu, J.~Yao, A.~Seff, K.~Cherry, L.~Kim, and R.~M.
  Summers, ``Improving computer-aided detection using convolutional neural
  networks and random view aggregation,'' \emph{IEEE Transactions on Medical
  Imaging}, vol.~35, no.~5, pp. 1170--1181, 2016.

\bibitem{moeskops2016automatic}
P.~Moeskops, M.~A. Viergever, A.~M. Mendrik, L.~S. de~Vries, M.~J. Benders, and
  I.~I{\v{s}}gum, ``Automatic segmentation of {MR} brain images with a
  convolutional neural network,'' \emph{IEEE Transactions on Medical Imaging},
  vol.~35, no.~5, pp. 1252--1261, 2016.

\bibitem{zhang2017deeppap}
L.~Zhang, L.~Lu, I.~Nogues, R.~Summers, S.~Liu, and J.~Yao, ``Deeppap: Deep
  convolutional networks for cervical cell classification,'' \emph{IEEE Journal
  of Biomedical and Health Informatics}, vol.~21, no.~6, pp. 1633--1643, 2017.

\bibitem{nie20163d}
D.~Nie, H.~Zhang, E.~Adeli, L.~Liu, and D.~Shen, ``3{D} deep learning for
  multi-modal imaging-guided survival time prediction of brain tumor
  patients,'' in \emph{MICCAI}.\hskip 1em plus 0.5em minus 0.4em\relax
  Springer, 2016, pp. 212--220.

\bibitem{yao2016imaging}
J.~Yao, S.~Wang, X.~Zhu, and J.~Huang, ``Imaging biomarker discovery for lung
  cancer survival prediction,'' in \emph{MICCAI}.\hskip 1em plus 0.5em minus
  0.4em\relax Springer, 2016, pp. 649--657.

\bibitem{zhu2017wsisa}
X.~Zhu, J.~Yao, F.~Zhu, and J.~Huang, ``{WSISA}: Making survival prediction
  from whole slide histopathological images,'' in \emph{CVPR}.\hskip 1em plus
  0.5em minus 0.4em\relax IEEE, 2017, pp. 7234--7242.

\bibitem{simonyan2014two}
K.~Simonyan and A.~Zisserman, ``Two-stream convolutional networks for action
  recognition in videos,'' in \emph{NIPS}, 2014, pp. 568--576.

\bibitem{feichtenhofer2016convolutional}
C.~Feichtenhofer, A.~Pinz, and A.~Zisserman, ``Convolutional two-stream network
  fusion for video action recognition,'' in \emph{CVPR}, 2016, pp. 1933--1941.

\bibitem{brox2004high}
T.~Brox, A.~Bruhn, N.~Papenberg, and J.~Weickert, ``High accuracy optical flow
  estimation based on a theory for warping,'' in \emph{ECCV}.\hskip 1em plus
  0.5em minus 0.4em\relax Springer, 2004, pp. 25--36.

\bibitem{graves2008supervised}
A.~Graves, ``Supervised sequence labelling with recurrent neural networks,''
  PhD Dissertation, Technical University of Munich, 2008.

\bibitem{ranzato2014video}
M.~Ranzato, A.~Szlam, J.~Bruna, M.~Mathieu, R.~Collobert, and S.~Chopra,
  ``Video (language) modeling: a baseline for generative models of natural
  videos,'' in \emph{arXiv}, 2014.

\bibitem{srivastava2015unsupervised}
N.~Srivastava, E.~Mansimov, and R.~Salakhudinov, ``Unsupervised learning of
  video representations using lstms,'' in \emph{ICML}, 2015, pp. 843--852.

\bibitem{finn2016unsupervised}
C.~Finn, I.~Goodfellow, and S.~Levine, ``Unsupervised learning for physical
  interaction through video prediction,'' in \emph{NIPS}, 2016, pp. 64--72.

\bibitem{mathieu2015deep}
M.~Mathieu, C.~Couprie, and Y.~LeCun, ``Deep multi-scale video prediction
  beyond mean square error,'' in \emph{ICLR}, 2016.

\bibitem{neverova2017predicting}
N.~Neverova, P.~Luc, C.~Couprie, J.~Verbeek, and Y.~LeCun, ``Predicting deeper
  into the future of semantic segmentation,'' in \emph{ICCV}, 2017.

\bibitem{bhattacharyya2016long}
A.~Bhattacharyya, M.~Malinowski, B.~Schiele, and M.~Fritz, ``Long-term image
  boundary extrapolation,'' in \emph{arXiv}, 2016.

\bibitem{maddison2014move}
C.~J. Maddison, A.~Huang, I.~Sutskever, and D.~Silver, ``Move evaluation in go
  using deep convolutional neural networks,'' in \emph{ICLR}, 2015.

\bibitem{silver2016mastering}
D.~Silver, A.~Huang, C.~J. Maddison, A.~Guez, L.~Sifre, G.~Van Den~Driessche,
  J.~Schrittwieser, I.~Antonoglou, V.~Panneershelvam, M.~Lanctot \emph{et~al.},
  ``Mastering the game of go with deep neural networks and tree search,''
  \emph{Nature}, vol. 529, no. 7587, pp. 484--489, 2016.

\bibitem{baker2011database}
S.~Baker, D.~Scharstein, J.~Lewis, S.~Roth, M.~J. Black, and R.~Szeliski, ``A
  database and evaluation methodology for optical flow,'' \emph{IJCV}, vol.~92,
  no.~1, pp. 1--31, 2011.

\bibitem{zhang2017personalized}
L.~Zhang, L.~Lu, R.~M. Summers, E.~Kebebew, and J.~Yao, ``Personalized
  pancreatic tumor growth prediction via group learning,'' in
  \emph{MICCAI}.\hskip 1em plus 0.5em minus 0.4em\relax Springer, 2017, pp.
  424--432.

\bibitem{bosman2010classification}
F.~T. Bosman, F.~Carneiro, R.~H. Hruban, N.~D. Theise \emph{et~al.}, \emph{WHO
  classification of tumours of the digestive system.}\hskip 1em plus 0.5em
  minus 0.4em\relax World Health Organization, 2010, no. Ed. 4.

\bibitem{neuroimage2006}
P.~A. Yushkevich, J.~Piven, H.~C. Hazlett, R.~G. Smith, S.~Ho, J.~C. Gee, and
  G.~Gerig, ``User-guided {3D} active contour segmentation of anatomical
  structures: significantly improved efficiency and reliability,''
  \emph{{NeuroImage}}, vol.~31, no.~3, pp. 1116--1128, 2006.

\bibitem{liu2009beyond}
C.~Liu, ``Beyond pixels: exploring new representations and applications for
  motion analysis,'' Ph.D. dissertation, Massachusetts Institute of Technology,
  2009.

\bibitem{wolfgang2013recent}
C.~L. Wolfgang, J.~M. Herman, D.~A. Laheru, A.~P. Klein, M.~A. Erdek, E.~K.
  Fishman, and R.~H. Hruban, ``Recent progress in pancreatic cancer,''
  \emph{CA: A Cancer Journal for Clinicians}, vol.~63, no.~5, pp. 318--348,
  2013.

\bibitem{yao2011everolimus}
J.~C. Yao, M.~H. Shah, T.~Ito, C.~L. Bohas, E.~M. Wolin, E.~Van~Cutsem, T.~J.
  Hobday, T.~Okusaka, J.~Capdevila, E.~G. De~Vries \emph{et~al.}, ``Everolimus
  for advanced pancreatic neuroendocrine tumors,'' \emph{New England Journal of
  Medicine}, vol. 364, no.~6, pp. 514--523, 2011.

\bibitem{jia2013caffe}
\BIBentryALTinterwordspacing
Y.~Jia, ``Caffe: An open source convolutional architecture for fast feature
  embedding,'' 2013. [Online]. Available:
  \url{http://caffe.berkeleyvision.org/}
\BIBentrySTDinterwordspacing

\end{thebibliography}
}

\end{document}